\documentclass{article}

\PassOptionsToPackage{numbers, sort&compress}{natbib}

\usepackage{natbib}
\usepackage{authblk}

\usepackage[utf8]{inputenc} %
\usepackage[T1]{fontenc}    %
\usepackage{lmodern}
\usepackage{hyperref}       %
\usepackage{url}            %
\usepackage{booktabs}       %
\usepackage{amsfonts}       %
\usepackage{nicefrac}       %
\usepackage{microtype}      %
\usepackage{xcolor}         %

\usepackage{amsmath}
\usepackage{amssymb}
\usepackage{mathtools}
\usepackage{amsthm}
\usepackage{caption}
\usepackage{wrapfig}

\usepackage{MnSymbol}

\DeclareMathOperator*{\argmax}{arg\,max}

\DeclareMathOperator*{\nn}{NN}

\usepackage{algorithm}
\usepackage{algpseudocode}
\usepackage{multirow}
\makeatletter
\renewcommand{\ALG@beginalgorithmic}{\small}
\makeatother

\usepackage[disable]{todonotes}

\newcommand{\todow}[1]{\todo[inline,color=cyan!20]{WL: #1}}

\title{Enhancing Label Sharing Efficiency in Complementary-Label Learning with Label Augmentation}

\author[1,2]{Wei-I Lin \footnote{\url{r10922076@csie.ntu.edu.tw}}}
\author[2]{Gang Niu \footnote{\url{gang.niu.ml@gmail.com}}}
\author[1]{Hsuan-Tien Lin \footnote{\url{htlin@csie.ntu.edu.tw}}}
\author[2,3]{Masashi Sugiyama \footnote{\url{sugi@k.u-tokyo.ac.jp}}}
\affil[1]{National Taiwan University}
\affil[2]{RIKEN}
\affil[3]{The University of Tokyo}
\date{}

\begin{document}

\maketitle

\begin{abstract}
Complementary-label Learning (CLL) is a form of weakly supervised learning that trains an ordinary classifier using only complementary labels, which are the classes that certain instances do not belong to. While existing CLL studies typically use novel loss functions or training techniques to solve this problem, few studies focus on how complementary labels collectively provide information to train the ordinary classifier. In this paper, we fill the gap by analyzing the implicit sharing of complementary labels on nearby instances during training. Our analysis reveals that the efficiency of implicit label sharing is closely related to the performance of existing CLL models. Based on this analysis, we propose a novel technique that enhances the sharing efficiency via complementary-label augmentation, which explicitly propagates additional complementary labels to each instance. We carefully design the augmentation process to enrich the data with new and accurate complementary labels, which provide CLL models with fresh and valuable information to enhance the sharing efficiency. We then verify our proposed technique by conducting thorough experiments on both synthetic and real-world datasets. Our results confirm that complementary-label augmentation can systematically improve empirical performance over state-of-the-art CLL models.
\end{abstract}

\section{Introduction}
Ordinary supervised learning relies on labeled data to train a classifier.
However, obtaining high-quality labels can be expensive or impractical in real-world scenarios.
To overcome the challenge, various types of weakly supervised learning (WSL) tasks have been studied in recent years which focus on learning from weaker information than high-quality ordinary labels.
Those includes but not limited to noisy-label learning, partial-label learning, and positive-unlabeled learning  \cite{zhou2018brief,sugiyama2022machine}.
Studies on WSL can potentially make a real-world classification task possible when only weak information is provided.

In this paper, we focus on complementary-label learning (CLL), a very weak type of WSL \cite{ishida2017learning}.
Complementary labels refer to the classes to which a given instance does not belong to.
CLL aims to learn from only complementary labels during training, while correctly predicting the ordinary labels of unseen instances.
Pioneering CLL researchers believe that studies on CLL could potentially make multi-class classification more realistic when ordinary labels are costly to obtain \cite{ishida2017learning,ishida2019complementary}.
Although there is no real-world dataset that demonstrates that CLL is a practical learning paradigm, studies on CLL still help understand the weakly supervised learning.
For instance, \citet{chou2020unbiased} revealed that the unbiased risk estimator, a popular method in weakly supervised learning, could be misleading due to the large variance in the gradient.
\citet{WL2023} found that CLL could be reduced to the problem of probability estimates.
\citet{deng2022boosting} utilized the information of pseudo complementary labels to improve the performance of semi-supervised learning.
These examples demonstrate the significance of fundamental studies on CLL.
\todow{I'm still unsure whether this revision is better than the original one. Could HT help check the paragraph again?}

\citet{ishida2017learning} initiated the study of CLL by rewriting the loss functions for ordinary classification with its unbiased risk estimate (URE) that depends only on complementary labels.
Despite the unbiasedness property, \citet{ishida2019complementary} and \citet{chou2020unbiased} discovered that the URE approach is prone to overfitting.
To conquer the URE's tendency to overfit, \citet{ishida2019complementary} argued that the issue arises because URE are not lower bounded.
They proposed two tricks, non-negative correction and gradient ascent, to effectively keep the URE loss non-negative during training.
\citet{chou2020unbiased} addressed the issue by proposing an alternative loss function, the surrogate complementary loss, that is bounded below by zero.
Despite its empirical effectiveness, \citet{chou2020unbiased} did not provide theoretical guarantees on the surrogate complementary loss.
To fill the gap, \citet{gao2021discriminative} modeled the posterior distribution of complementary labels, leading to a non-negative loss function, and proved the statistical consistency of the estimator.
\citet{liuconsistent} proposed the order-preserving loss, enabling training from a family of loss functions on the CLL problem with statistical consistency guarantees while preventing the overfitting issue.
The analysis or the benchmark experiments of the methods mentioned above, however, rely on the assumption that the complementary labels are generated uniformly.
To accommodate biased complementary-label generation, \citet{yu2018learning} proposed a forward-correction loss that directly uses the transition matrix to model the non-uniform generation.
This allows learning from CLs with biased generation.

In this paper, we first find that the success of the loss-based methods for CLL relies on the \emph{implicit} shares of complementary labels through the smoothness of neural networks.
Specifically, we observe a strong correlation between the efficiency of the implicit label sharing and the model performance.
Based on the result, we propose a novel technique that explicitly shares the complementary labels between neighboring instances.
The proposed technique provides two advantages.
First, it is compatible with most previous approaches on CLL.
As the proposed complementary-label augmentation generates a soft complementary label for each instance, it becomes compatible with any CLL algorithm that can take soft complementary labels as input.
Second, the compatibility suggests that the proposed method has the potential to provide conceptually orthogonal benefits to the existing methods.
We confirm through the experiments on the real-world and synthetic complementary datasets that the proposed complementary-label augmentation does enhance the existing CLL algorithms.
Our contribution can be summarized as follows.
\begin{enumerate}
    \item We proposed \emph{complementary-label augmentation}, a technique that explicitly shares the complementary labels between neighboring instances before training. In addition, the method is compatible with the existing CLL algorithms and can provide conceptually orthogonal benefits to them.
    \item The proposed method is based on our empirical observation that (a) the success of the previous loss-based CLL algorithms can be attributed to the implicit shares of the labeling information between data instances, (b) a strong relationship between the implicit label sharing efficiency and model performance, and (c) explicitly sharing the complementary labels can enhance the sharing efficiency, leading to improved classification accuracy.
    \item Extensive experiments on both real-world and synthetic complementary datasets confirm that the proposed method improves the existing loss-based CLL algorithms.
\end{enumerate}

\section{Problem Setting}
In this section, we first introduce the problem of Complementary-Label Learning (CLL) in Section~\ref{subsec:complementary-label-learning}, and then discuss some common assumptions on how the complementary labels are generated in Section~\ref{subsec:generateion-of-cl}.

\subsection{Complementary-Label Learning}
\label{subsec:complementary-label-learning}
CLL is a weakly-supervised learning problem on multi-class classification. 
Typical multi-class classification assumes that the feature vector $x_i$'s and the corresponding labels $y_i$'s in the training dataset $D=\{(x_i, y_i)\}_{i=1}^N$ are i.i.d. sampled from an unknown distribution.
In CLL, $D$ is not accessible to the learning algorithm.
Instead, a complementary dataset $\bar{D}=\{(x_i, \bar{y}_i)\}_{i=1}^N$ is provided to the learner, where $\bar{y}_i$ denotes a complementary label, a class to which the instance $x_i$ does not belong.
The goal of the complementary learning algorithm is to find a classifier $f$ that can predict unseen instances correctly.
Typically, classifier $f$ is implemented by a decision function $g: \mathbb{R}^d\to \mathbb{R}^K$ and taking argmax on $g$, i.e., $f(x) = \argmax_{k\in[K]} g(x)_k$, where $g(x)_k$ denote the $k$-th element of $g(x)$ and $[K]=\{1,\dotsc,K\}$ denote the set of labels.

\subsection{Generation of Complementary Labels}
\label{subsec:generateion-of-cl}
In the CLL literature, some assumptions on the generation of complementary labels (CL) are made. The most simple one is called \emph{uniform generation}. Firstly proposed in the pioneering work \cite{ishida2017learning}, it assumes that each complementary label $\bar{y}_i$ is independently and uniformly selected from all the labels except the correct one. This assumption is also utilized in some subsequent works \cite{ishida2019complementary,chou2020unbiased}.

A further generalization to the uniform case is called class-conditional assumption. It assumes that the generation process of the CLs depends only on their underlying ordinary labels, i.e., there is $T_{ij} = P(\bar{y}=j\mid y=i)$ for each ordinary label $i$ and complementary label $j$. When $T_{ij} = \tfrac{1}{K-1}$ for each $i\neq j$, this falls back to the uniform generation. CLL under such assumption is further analyzed in \cite{yu2018learning}. This type of generation is also called \emph{biased generation} when $T_{ij}$ is not always $\frac{1}{K-1}$ whenever $i\neq j$ to distinguish it from the uniform one.
The generation process mentioned above is noiseless, meaning there are no CLs that  actually belong to the ordinary class.
In the noiseless case, $T_{ii} = 0$ holds for each $i\in [K]$, whereas in noisy CLL \cite{ishiguro2022learning}, the diagonal elements of the transition matrix $T_{ii}$ are a small positive number.

In our work, the augmented CLs using the proposed method are not generated with respect to the class-conditional assumption.
Nevertheless, we will show in the experimental section that the proposed method still improves the learning algorithms that rely on them.

\section{Proposed Method}\label{sec:methods}
In this section, we first establish the relationship between the implicit label sharing and model performance for the CLL algorithms in Section~\ref{subsec:implicit-share}.
Then, we demonstrate that the implicit sharing efficiency could be enhanced by explicitly sharing the complementary labels through the illustrative experiments in Section~\ref{subsec:explicit-share}.
Finally, we describe the proposed \emph{label augmentation}, a practical way to explicitly share the complementary labels, and discuss some crucial components when performing label augmentation in Section~\ref{subsec:complementary-label-augmentation}.

\subsection{Implicit Label Sharing and Model Performance}
\label{subsec:implicit-share}

\paragraph{Implicit Label Sharing}
We first take a closer look at the training process of the CLL algorithms.
Let us consider a $K$-class CLL problem with a single complementary label per instance.
Intuitively, if a model is able to recognize only one complementary label per instance, then the accuracy for the model is at best $\frac{1}{K-1}$ by randomly guessing from the remaining labels.
However, previous CLL methods can attain much higher accuracies than $\frac{1}{K-1}$ \cite{ishida2017learning,ishida2019complementary,chou2020unbiased,gao2021discriminative,WL2023,liuconsistent}.
That leads us to conjecture that one of the mechanisms behind the current CLL algorithms is the implicit sharing of the complementary labels through the smoothness of the neural networks.
Intuitively, if two instances are near in the feature space learned by the neural network, then they will have similar probability outputs.
This smoothness can potentially let the complementary labels be shared implicitly to their neighbors in the feature space.

To verify the conjecture, we performed a toy experiment. First, we divided the labels into three types:
(a) Seen complementary labels, which are the labels provided in the complementary dataset
(b) Ordinary labels, and
(c) Unseen complementary labels, which are the remaining labels. These labels are the complementary labels that are not provided to the learning algorithms.
Then, we investigated how much confidence the model allocates to the seen and unseen complementary labels during training.
Specifically, for a model $f$ with a probabilistic output, we checked (a) $f_{\bar{y}}(x)$ for seen complementary labels and (b) $\frac{1}{K-2}\sum_{y'\notin\{y,\bar{y}\}} f_{y'}(x)$ for unseen complementary labels.
We trained the models with the SCL-NL loss function \cite{chou2020unbiased} and reported the results in Figure~\ref{fig:toy-1}. Other experimental details are left in Appendix~\ref{subsec:appendix-toy-details}.

We drew the following observations from Figure~\ref{fig:toy-1}.
First, the blue lines, indicating the mean confidence on the given complementary labels, are minimized to zero in all the datasets. This implies that the learning algorithm is able to memorize and fit the given complementary labels perfectly.
Second, the green lines, indicating the mean confidence on the unseen complementary labels, become smaller but do not reach zero in all the datasets.
This phenomenon suggests that somehow the learning algorithm implicitly shares the information of the given complementary labels to other data instances; otherwise, the confidence on the unseen complementary labels will remain unchanged during the whole training process.

\begin{figure}[t]
    \vskip-0.15in
    \centering
    \resizebox{\columnwidth}{!}{
    \includegraphics[width=0.3\linewidth]{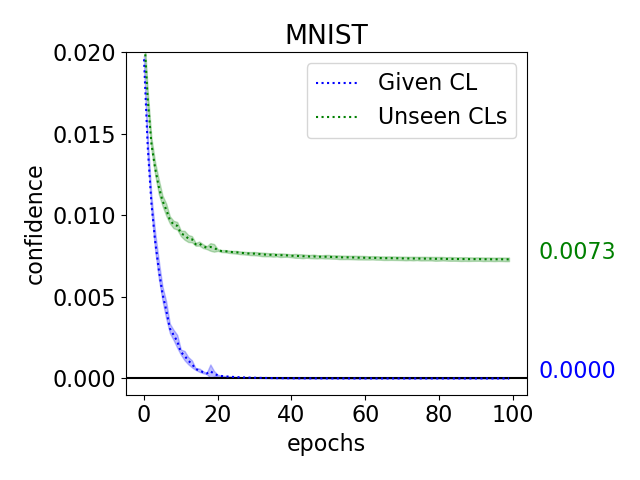}
    \includegraphics[width=0.3\linewidth]{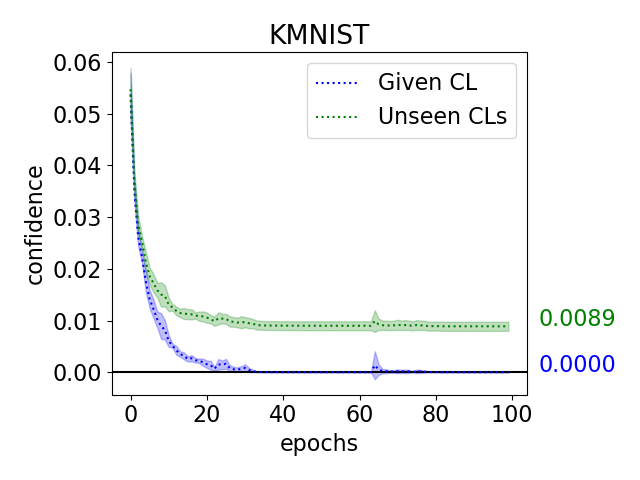}
    \includegraphics[width=0.3\linewidth]{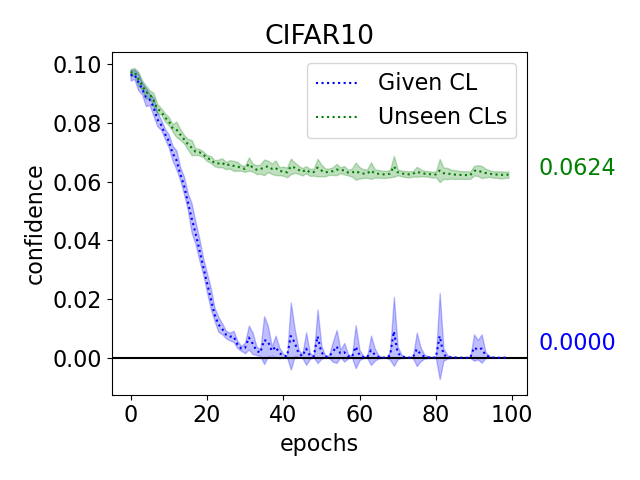}
    }
    \caption{
        Comparison of the complementary label confidence during training with the original complementary datasets.
        The black horizontal lines indicate the zero confidence.
        The colored numbers indicate the corresponding confidences in the last epoch.
    }
    \label{fig:toy-1}
\end{figure}
\begin{figure}[t]
    \vskip-0.15in
    \centering
    \resizebox{\columnwidth}{!}{
    \includegraphics[width=0.3\linewidth]{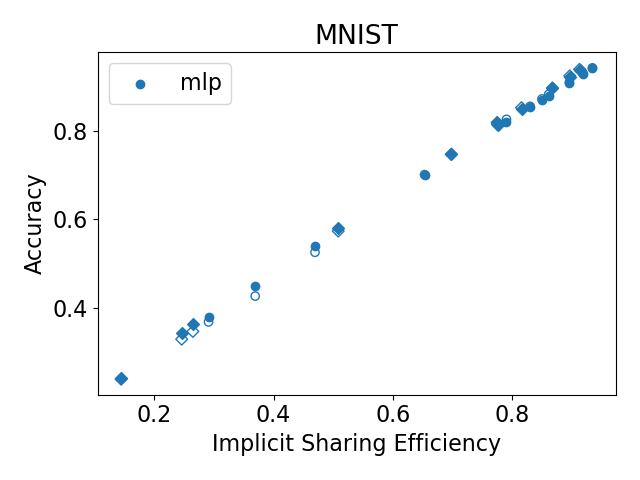}
    \includegraphics[width=0.3\linewidth]{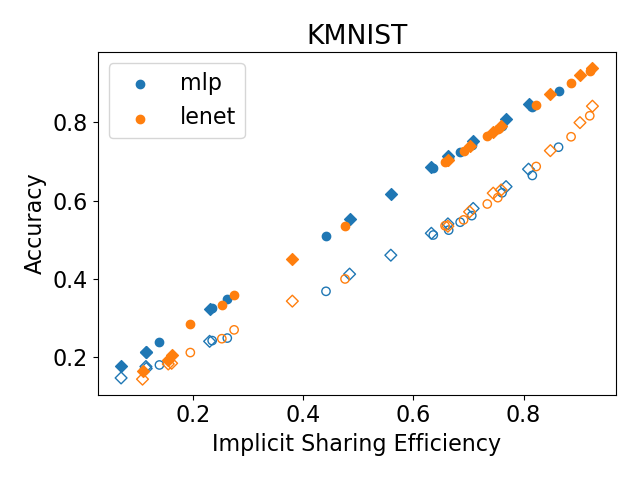}
    \includegraphics[width=0.3\linewidth]{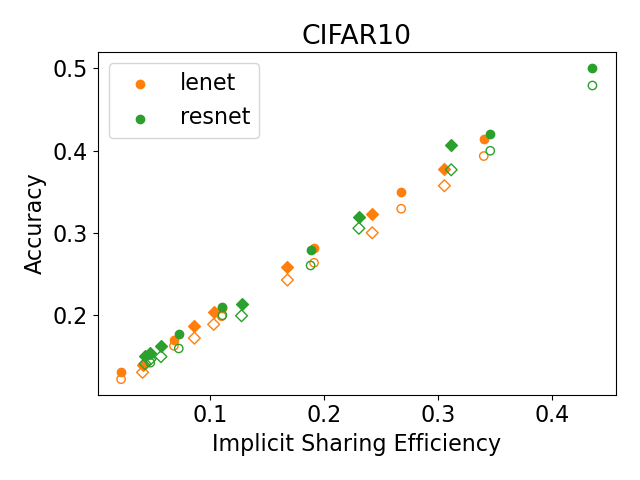}
    }
    \caption{
        Relationship between the implicit sharing efficiency and model performance.
        Each dot represents a result with certain dataset size and model architecture.
        Solid markers ($\bullet/\filleddiamond$) refer to the training accuracy while empty markers ($\circ/\diamond$) refer to the testing accuracy.
        Circles ($\bullet/\circ$) and diamonds ($\filleddiamond/\diamond$) refer to the experiments with the AdamW and the SGD optimizers, respetively.
    }
    \label{fig:toy-2}
\end{figure}
\paragraph{Relationship between Implicit Label Sharing and Performance}
Now that we confirmed that the implicit sharing occurs during the training process of CLL algorithms, we wonder whether the \emph{efficiency} of the implicit sharing is related to the model performance. To clarify their relationship, we first define the implicit sharing efficiency as follows:
\begin{equation*}
    \text{Implicit sharing efficiency}
    = 1 - \frac{1}{N}\sum_{i=1}^N \frac{K-1}{K-2}\sum_{\bar{y}'\notin\{y_i,\bar{y}_i\}} f_{\bar{y}'}(x_i).
\end{equation*}
This metric measures the magnitude of the reduction in the model's confidence on the unseen CLs.
If the implicit sharing helps identify all the complementary labels, then $f_{\bar{y}'}(x_i)$ will become zero, making the implicit sharing efficiency becomes one. 
If there is no implicit sharing between the instances, then $f_{\bar{y}'}(x_i)$ will become $\frac{1}{K-1}$ on average. In this case, implicit sharing efficiency becomes zero.
For instance, in the MNIST experiment in Figure~\ref{fig:toy-1}, if there is no implicit label sharing, the mean confidence on unseen CLs is $1/9$.
We observe that empirically the mean confidence on unseen CLs after training is actually $0.0073$, suggesting that the mean confidence on unseen CLs reduce by $1-0.0073/\tfrac{1}{9}$, which is 93.43\%.
Hence, the implicit sharing efficiency is 93.43\% in this case.

To vary the implicit sharing efficiency, we performed the same experiment with different dataset sizes, model architectures, and optimizers.
We reported the relationship between the implicit sharing efficiency and the model performance in Figure~\ref{fig:toy-2}

As we can observe from the figure, there is a near-linear relationship between the implicit sharing efficiency and the model's training accuracy.
This relationship is strong and independent of the network architectures, dataset size, and optimizers.
The higher the implicit sharing is, the higher the model's training accuracy is.
The improved training accuracy indicates that the model becomes better at recognizing the true labels in the training dataset, leading to better testing accuracy.
The result highlights the importance of improving the implicit sharing efficiency when designing algorithms for CLL.
We also performed experiments with other CLL algorithms and reported the results in Appendix~\ref{subsec:appendix-toy-additional-results}.
This relationship also holds for other CLL algorithms.

\subsection{Illustrative Experiments on Explicit Label Sharing}
\label{subsec:explicit-share}

\begin{table}[t]
\vskip-0.15in
\centering

\resizebox{\columnwidth}{!}{
\includegraphics[width=0.3\linewidth]{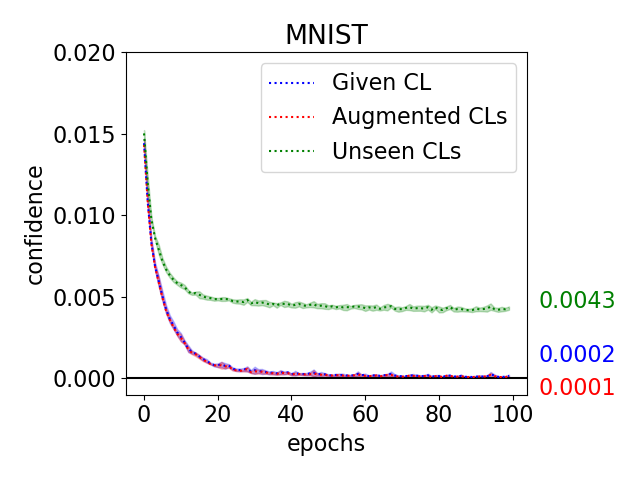}
\includegraphics[width=0.3\linewidth]{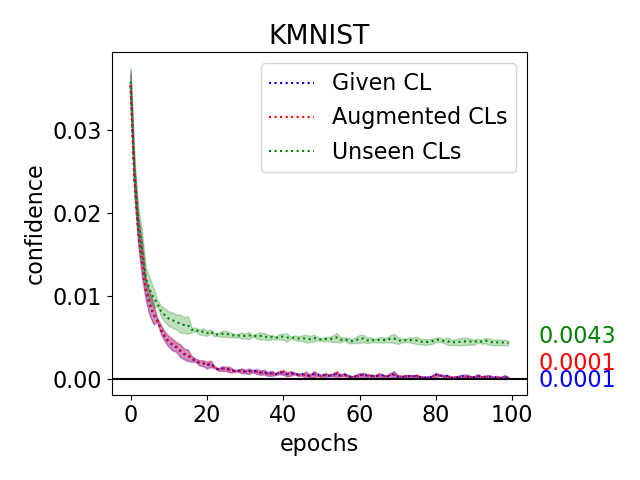}
\includegraphics[width=0.3\linewidth]{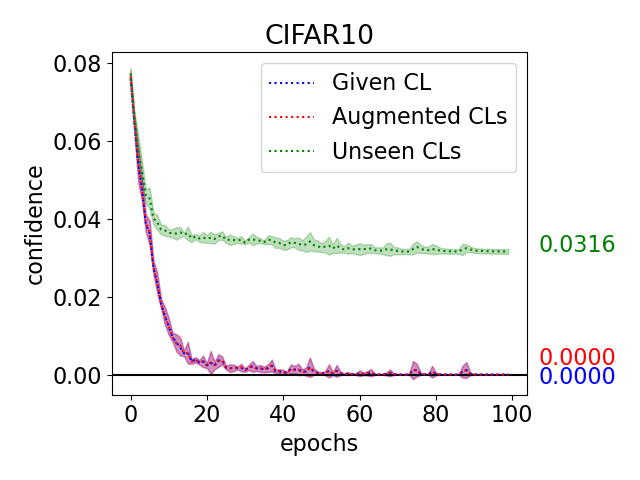}
}
\captionof{figure}{
    Comparison of the complementary label confidence during training with the explicit shares of complementary labels.
}
\label{fig:toy}

\captionof{table}{
The performance (in \%) and noise rate (in \%) on the illustrative toy experiments with different datasets and with different settings: original, or with explicit shares.
The numbers before the arrow is from the original setting whereas the numbers after the arrow are from the setting with explicit shares.
(original $\Rightarrow$ with explicit shares)
}
\label{table:toy}

\begin{scriptsize}
\begin{tabular}{lccc}
\toprule
Dataset     & Training Accuracy    & Testing Accuracy & Neighboring Noise Rate \\
\midrule
MNIST    & 94.24 $\pm$ 0.09 $\Rightarrow$ 98.32 $\pm$ 0.06
    & 94.34 $\pm$ 0.21 $\Rightarrow$ 97.26 $\pm$ 0.07
    & 0.73 $\pm$ 0.02 $\Rightarrow$ 0.64 $\pm$ 0.02 \\
KMNIST   & 92.95 $\pm$ 0.99 $\Rightarrow$ 98.09 $\pm$ 0.17
    & 81.63 $\pm$ 0.89 $\Rightarrow$ 91.38 $\pm$ 0.65
    & 0.73 $\pm$ 0.08 $\Rightarrow$ 0.28 $\pm$ 0.02 \\
CIFAR10  & 50.07 $\pm$ 1.79 $\Rightarrow$ 86.96 $\pm$ 0.32
    & 47.93 $\pm$ 1.90 $\Rightarrow$ 78.07 $\pm$ 1.68
    & 4.51 $\pm$ 0.20 $\Rightarrow$ 1.21 $\pm$ 0.13 \\
\bottomrule
\end{tabular}
\end{scriptsize}
\end{table}
\paragraph{Explicit Complementary Label Sharing}
Although the previous CLL algorithms are able to implicitly share complementary labels, there is still room for improvement in the sharing efficiency.
To analyze why the previous CLL algorithms could not attain perfect implicit label sharing, we inspected the learned representation space at the end of training.
To be specific, we checked the noise rate of the complementary labels from the neighboring 128 instances on the learned representation space.
A complementary label from the neighboring instances is considered noisy if it is actually the ordinary label.
The results are reported in Table~\ref{table:toy}.
As we can see, the complementary labels from the neighboring instances are not always correct.
The phenomenon of the incorrect label sharing may explain why the implicit sharing efficiency could not attain one.

Based on the results, we wonder whether it is possible to learn a better representation space and thereby improve the implicit sharing efficiency through \emph{explicitly} sharing complementary labels.
To answer the question, for each data instance, we first randomly added half of the unseen complementary labels to the complementary dataset, and then trained a model using both the original complementary labels and the augmented ones.
The rest of the settings is the same as the previous experiment, and we reported the results in Figure~\ref{fig:toy} and Table~\ref{fig:toy}.

From the results, we observe that the benefits of explicitly adding more complementary labels are three-fold.
First, the mean confidence on the augmented complementary labels also goes down to near zero at the end of the training, suggesting that the learning algorithm is able to memorize more complementary labels.
Second, we see a reduction in the neighboring noise rate in the learned representation space, indicating that the explicit sharing helps learn a less noisy representation space.
Third, the mean confidence on the unseen complementary labels is smaller in the experiments with the explicit sharing than the original experiments without the explicit label sharing.
This phenomenon suggests that the explicit sharing helps improve the implicit label sharing efficiency.
This efficiency improvement leads to improved accuracies on both the training and testing datasets, as shown in Table~\ref{table:toy}.

\paragraph{Challenges in Explicitly Sharing Complementary Labels}
Although the previous experiments demonstrated the potential benefits of explicitly adding more complementary labels, it remains challenging in practice to do so without incurring extra label collection costs.
Besides, obtaining new complementary labels from the existing dataset may introduce noise to the complementary dataset.
How to share informative complementary labels while keeping the noise rate low becomes the main challenge in the explicit label sharing.

\subsection{Complementary-Label Augmentation}
\label{subsec:complementary-label-augmentation}

\begin{table*}[t]

\centering
\begin{algorithm}[H]
\captionof{algorithm}{Complementary-Label Augmentation}
\label{alg:algo}
\begin{algorithmic}[1]
   \State {\bfseries Input:} Complementary Labels $\bar{Y}$, Affinity Matrix $W$, Number of training samples $N$, Hyperparameters $T$, $\alpha$
   \State {\bfseries Output:} Augmented Complementary Labels $Z$
   \State $Z = \bar{Y}$
   \For{$i=1$ {\bfseries to} $N$}
   \State $W_i = W_i / \sum_{j=1}^N W_{i,j}$ \Comment{normalizing the weights}
   \EndFor
   \For{$t=1$ {\bfseries to} $T$}
   \State $Z = \alpha \bar{Y} + (1-\alpha)WZ$
   \EndFor
   \For{$i=1$ {\bfseries to} $N$}
   \State $Z_i = Z_i / \sum_{j=1}^N Z_{i,j}$ \Comment{normalizing the soft labels}
   \EndFor
\end{algorithmic}
\end{algorithm}

\captionof{table}{
A summary of different label augmentation schemes.
}
\label{table:summary-conf}
\begin{scriptsize}
\begin{tabular}{lcc}
\toprule
Schemes & Affinity Matrix $W$ & Round $T$ \\
\midrule
Traditional & None & 0 \\
\midrule
\multirow{2}{*}{Rank-weighted Single Step} & \multirow{3}{*}{$W_{i,j} = \begin{cases*}
    \tfrac{1}{k} &\text{if $v_j$ is $v_i$'s $k$th neighbor ($k\leq N_K$)} \\
    0 &\text{otherwise}
\end{cases*}$} & \multirow{2}{*}{1} \\
\multirow{2}{*}{Rank-weighted Multi Steps} & & \multirow{2}{*}{100} \\
\\
\midrule
\multirow{2}{*}{Distance-weighted Single Step} & \multirow{3}{*}{$W_{i,j} = \begin{cases*}
    \exp(-\gamma\lVert v_i-v_j\rVert^2) &\text{if $v_j\in \nn_{N_K}(v_i)$} \\
    0 &\text{otherwise}
\end{cases*}$} & \multirow{2}{*}{1} \\
\multirow{2}{*}{Distance-weighted Multi Steps} & & \multirow{2}{*}{100} \\
\\
\bottomrule
\end{tabular}
\end{scriptsize}
\end{table*}
In this section, we proposed \emph{complementary-label augmentation} for the existing loss-based complementary learning algorithms.
The main idea of the complementary-label augmentation is to add new complementary labels from a data instance's neighbors.
The idea utilizes the smoothness of the representation space, i.e., if two instances have similar features, then they are likely to belong to the same class. That implies that the complementary labels from the neighboring instances can be shared to each other.

To do so, given a feature extractor $v(\cdot)$, let $v_i = v(x_i)$ denote the feature of the $i$th instance in the complementary dataset, and let $V=\{v_i\}_{i=1}^N$ denote the collection of extracted features.
For each data instance $x_i$ in $\{x_i\}_{i=1}^N$, let $\bar{y}_{i,k}$ denote the complementary label from $v_i$'s $k$th nearest neighbor in $V\backslash\{v_i\}$.

In a loss-based complementary learning algorithm, there is a loss function $\ell:[K]\times\mathbb{R}^K\to\mathbb{R}$ that takes a complementary label $\bar{y}_i$ and the model's prediction $g(x_i)$ as input. The learning algorithm minimizes the loss function $\ell$ with respect to the complementary dataset $\bar{D}$, i.e., the learning algorithm optimizes $g$ with respect to the following empirical risk:
\begin{equation}
    R(g;\ell) = \frac{1}{N} \sum_{i=1}^N \ell(\bar{y}_i, g(x_i)).
\end{equation}
A naive way to utilize the augmented labels is to directly add them to the loss function as follows:
\begin{equation}
    R'(g;\ell) = \frac{\alpha}{N} \sum_{i=1}^N \ell(\bar{y}_i, g(x_i)) + \frac{1-\alpha}{N} \sum_{i=1}^N \sum_{k=1}^{N_K} \ell(\bar{y}_{i,k}, g(x_i)),
\end{equation}
where $\alpha$ is a hyperparameter that controls how much weight to put on the augmented labels and $N_K$ is the hyperparameter that denotes the number of neighbors to consider.
This simple approach, however, neglects the fact that the complementary labels augmented from the neighboring instances are noisy.
To overcome the issue, traditional $k$NN methods typically associate a weight to the neighbors \cite{dudani1976distance}.
We follow the idea, and propose to use a weight $w_{i,k}$ for the pair $(x_i, \bar{y}_{i,k})$ when training with a complementary learning algorithm.
The loss function then becomes:
\begin{equation}
    R'(g;\ell) = \frac{\alpha}{N} \sum_{i=1}^N \ell(\bar{y}_i, g(x_i)) + \frac{1-\alpha}{N} \sum_{i=1}^N \sum_{k=1}^{N_K} w_{i,k} \ell(\bar{y}_{i,k}, g(x_i)).
\end{equation}
The above loss function can be interpreted as training with respect to soft complementary labels as follows. Let $z_i = \alpha e_{\bar{y}_i} + (1-\alpha) \sum_{i=1}^{N_k} w_{i,k} e_{\bar{y}_{i,k}}$, where $e_k$ denotes the one-hot vector of label $k$, then the loss function is equivalent to
\begin{equation}\label{eq:loss-func}
    R_\text{LA}(g;\ell) = \frac{1}{N} \sum_{i=1}^N \sum_{k=1}^K z_{i,k} \ell(k, g(x_i)).
\end{equation}
To simplify the notations, we use an $N\times N$ matrix $W$ to denote the weight, where $W_{i,j}$ is the weight from instance $j$ to instance $i$.
In addition, we define $N\times K$ matrices $Y$ and $Z$ by setting the $i$th row of $Y$ to the one-hot vector of $\bar{y}_i$, i.e., $Y_i = e_{\bar{y}_i}$ and setting the $i$th row of $Z$ to $z_i$, i.e., $Z_i = z_i$, for each $i$.
Then, the process of complementary-label augmentation can be simplified to calculating the matrix $Z$ with the following equation: $Z = \alpha \bar{Y} + (1-\alpha)W\bar{Y}$, then optimize the model $g$ with respect to the loss function $R_\text{LA}$ defined in Equation~\ref{eq:loss-func}.

\paragraph{Weight and multi-step augmentation}
Intuitively, the complementary labels from far neighbors are noisier than the ones from near neighbors.
To ensure that the augmented complementary labels are not dominated by the noisy ones, the weights on the farther neighbors should be smaller than the weights on the nearer ones.
As a result, we consider two ways to set the affinity matrix $W$:
\begin{enumerate}
    \item
    \textbf{Rank-based approach}: weight based on the rank of the neighbors. Specifically, $W_{i,j} = \frac{1}{k}$ if $v_j$ is $v_i$'s $k$th neighbor with $k\leq N_K$.
    \item
    \textbf{Distance-based approach}: weight based on the distance from the neighbors. Specifically, $W_{i,j} = \exp(-\gamma\lVert v_i-v_j\rVert^2)$ if $v_j$ is within the $N_K$th nearest neighbor of $v_i$.
\end{enumerate}

The augmentation process proposed above can be performed multiple times.
By augmenting the labels multiple times, the label information could potentially be shared to more distant neighbors.
We call this technique multi-step label augmentation.
The pseudocode for the label augmentation is presented in Algorithm~\ref{alg:algo}, where we use $T$ to control the number of times to perform the complementary-label augmentation.
Different affinity matrix $W$ and the number of augmentation $T$ produce different augmented complementary labels.
We summarized in Table~\ref{table:summary-conf} the different schemes to perform the augmentation.

\subsection{Relation to other methods}
Label Propagation \cite{zhu2002learning,zhou2003learning} is a method in semi-supervised learning that also leverages the smoothness assumption.
The proposed multi-step complementary-label augmentation shares a similar form to the label propagation method; however, complementary-label augmentation is different from label propagation in two aspects.
First, label propagation in semi-supervised learning and complementary-label augmentation propagates different labeling information and for different goals.
The former propagates the \emph{ordinary} labels to the unlabeled data with the goal of obtaining pseudo labels for the unlabeled instances.
On the other hand, the latter propagates the \emph{complementary} labels to the neighboring instances in order to enrich the complementary labels of all the instances in the dataset.
Second, the two methods are different in their roles in the training process.
Label propagation is typically part of a training process \cite{iscen2019label}, where the labels and the model parameters are updated in an alternating manner.
In contrast, complementary-label augmentation is a technique to enrich the labeling information that is independent of the training process and is applied before the whole training process.

\section{Experiments}\label{sec:experiments}
To verify the efficacy of the proposed method, we conducted experiments on various benchmarks, including some synthetic datasets, where the complementary labels were generated uniformly, and some real-world datasets, where the complementary labels were annotated by humans.
The synthetic datasets in the benchmark include CIFAR10 and CIFAR20.
Both datasets contain 50,000 training samples and 10,000 testing samples, while CIFAR10 contains 10 classes and CIFAR20 contains 20 superclasses from CIFAR100.
We consider the setting of a single complementary label per data instance, where the complementary labels were generated uniformly.
We do not benchmark on CIFAR100 as we found that no previous CLL algorithms can learn a meaningful classifier in this dataset, provided only one complementary label per data instance, even with the proposed label augmentation.
The real-world datasets in the benchmark include CLCIFAR10 and CLCIFAR20, which contain the images in CIFAR10 and CIFAR20, respectively.
Each image in the datasets is annotated with three complementary labels by different human annotators.

\paragraph{Baseline Methods}
Four SOTA methods were considered in our experiments:
(a) PC \cite{ishida2017learning}: the pairwise comparison loss,
(b) URE-GA \cite{ishida2019complementary}: the unbiased risk estimator on cross-entropy loss with the gradient ascent trick,
(c) SCL-NL \cite{chou2020unbiased}: the surrogate complementary loss with the negative log loss, and
(d) L-W \cite{gao2021discriminative}: the weighted loss derived from the discriminative modeling of the complementary labels.
We did not include the forward correction method \cite{yu2018learning} as it is equivalent to SCL-NL when the transition matrix is uniform.
Each baseline method was trained directly without label augmentation, and trained with four configurations of label augmentation as in Table~\ref{table:summary-conf}.
Other implementation details, including hyperparameter selection through a validation process, are left in Appendix~\ref{subsec:appendix-exp-details}.

\begin{table}[t]
\caption{
Effects of label augmentation applied on different complementary learning algorithms on CIFAR10 and CIFAR20. The mean and standard deviation of the testing accuracies over five random trials are reported. RSS, RMS, DSS and DMS refer to rank-weighted single-step, rank-weighted multi-step, distance-weighted single-step and distance-weighted multi-step, respectively.
}
\label{table:clean-bench}
\centering
\begin{scriptsize}
\begin{tabular}{lcccc}
\toprule
        & CIFAR10 & CIFAR20 & CLCIFAR10 & CLCIFAR20 \\
\midrule
PC     &35.59 $\pm$ 0.44 &11.01 $\pm$ 1.79 &40.69 $\pm$ 1.98 &13.79 $\pm$ 0.38 \\
PC+RSS &73.50 $\pm$ 0.54 &34.11 $\pm$ 0.46 &65.47 $\pm$ 0.61 &19.30 $\pm$ 2.12 \\
PC+RMS &83.47 $\pm$ 0.17 &55.00 $\pm$ 0.26 &74.83 $\pm$ 0.80 &23.26 $\pm$ 1.39 \\
PC+DSS &80.86 $\pm$ 0.25 &43.38 $\pm$ 0.50 &72.10 $\pm$ 0.54 &22.05 $\pm$ 0.44 \\
PC+DMS &\textbf{83.90 $\pm$ 0.25} &\textbf{57.62 $\pm$ 0.49} &\textbf{75.13 $\pm$ 0.49} &\textbf{26.43 $\pm$ 2.02} \\
\midrule
URE-GA     &58.23 $\pm$ 1.45 &12.03 $\pm$ 0.49 &23.98 $\pm$ 2.91 &9.00 $\pm$ 0.38  \\
URE-GA+RSS &72.72 $\pm$ 1.83 &23.93 $\pm$ 3.25 &\textbf{28.66 $\pm$ 1.17} &11.52 $\pm$ 0.51 \\
URE-GA+RMS &73.33 $\pm$ 2.60 &25.36 $\pm$ 2.12 &25.03 $\pm$ 1.35 &12.32 $\pm$ 0.14 \\
URE-GA+DSS &72.97 $\pm$ 1.89 &27.13 $\pm$ 2.74 &25.42 $\pm$ 0.31 &12.49 $\pm$ 0.53 \\
URE-GA+DMS &\textbf{75.06 $\pm$ 3.01} &\textbf{29.73 $\pm$ 3.09} &20.76 $\pm$ 1.66 &\textbf{13.19 $\pm$ 0.30} \\
\midrule
SCL-NL     &73.95 $\pm$ 0.86&23.24 $\pm$ 1.54&44.12 $\pm$ 1.49 &8.09 $\pm$ 0.13  \\
SCL-NL+RSS &85.77 $\pm$ 0.78&54.71 $\pm$ 2.05&56.52 $\pm$ 0.54 &14.33 $\pm$ 1.20 \\
SCL-NL+RMS &88.67 $\pm$ 0.09&64.48 $\pm$ 0.58&73.30 $\pm$ 2.12 &19.74 $\pm$ 1.66 \\
SCL-NL+DSS &87.00 $\pm$ 0.32&55.08 $\pm$ 1.51&67.86 $\pm$ 1.87 &17.19 $\pm$ 1.51  \\
SCL-NL+DMS &\textbf{88.72 $\pm$ 0.15} &\textbf{66.33 $\pm$ 0.42} &\textbf{74.87 $\pm$ 2.25} &\textbf{24.44 $\pm$ 2.88} \\
\midrule
L-W     &50.26 $\pm$ 0.37 &15.64 $\pm$ 2.85 &38.38 $\pm$ 0.63 &7.35 $\pm$ 0.80 \\
L-W+RSS &83.13 $\pm$ 0.22 &42.35 $\pm$ 2.67 &54.58 $\pm$ 0.65 &13.18 $\pm$ 0.90 \\
L-W+RMS &88.27 $\pm$ 0.24 &63.28 $\pm$ 0.51 &73.25 $\pm$ 0.61 &20.05 $\pm$ 2.49 \\
L-W+DSS &86.45 $\pm$ 0.25 &47.43 $\pm$ 1.57 &67.84 $\pm$ 0.62 &17.49 $\pm$ 0.69 \\
L-W+DMS &\textbf{88.45 $\pm$ 0.24} &\textbf{65.68 $\pm$ 0.24} &\textbf{74.76 $\pm$ 0.68} &\textbf{22.09 $\pm$ 1.86} \\
\bottomrule
\end{tabular}
\end{scriptsize}
\end{table}

\paragraph{Results and Discussion}
The benchmark results are reported in Table~\ref{table:clean-bench}.
As shown in the table, complementary-label augmentation improved all the baseline methods on all the datasets.
Taking a closer look, we found a general trend that the distance-based weighting performed better than the rank-based weighting.
The reason could be that the rank-based weighting could not distinguish between a high-density neighborhood and low-density neighborhood.
In low-density region, the first few neighbors could already be noisy, but rank-based weighting did not take this into consideration and reduce their weights.
In contrast, the weight-based weighting reduced their weights in this case.
On the other hand, another observation from Table~\ref{table:clean-bench} was that the multi-step augmentation outperformed the single-step augmentation.
This echoes our previous conjecture that multi-step augmentation could lead to better performance by further propagating the information of complementary labels to more distant instances.

\section{Ablation and Further Discussions}

\subsection{Effects on the Number of Neighbors}
\begin{figure}
    \vskip-0.15in
    \centering
    \resizebox{\columnwidth}{!}{
    \includegraphics[width=0.24\linewidth]{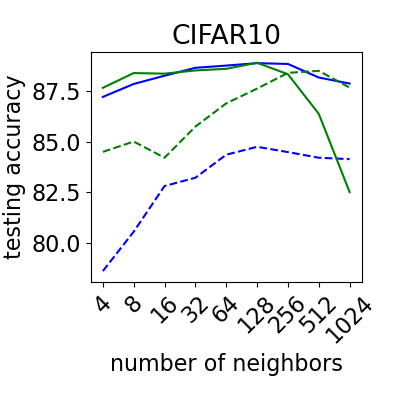}
    \includegraphics[width=0.24\linewidth]{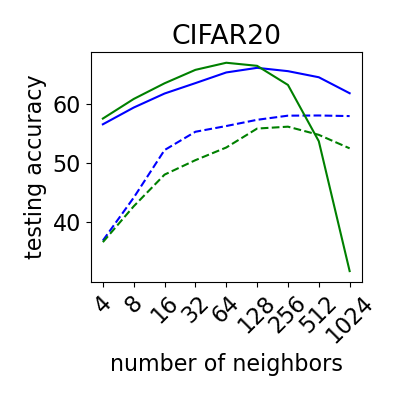}
    \includegraphics[width=0.24\linewidth]{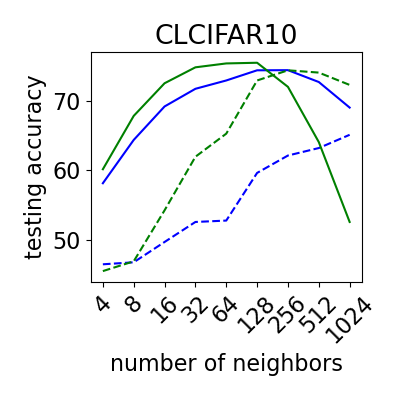}
    \includegraphics[width=0.24\linewidth]{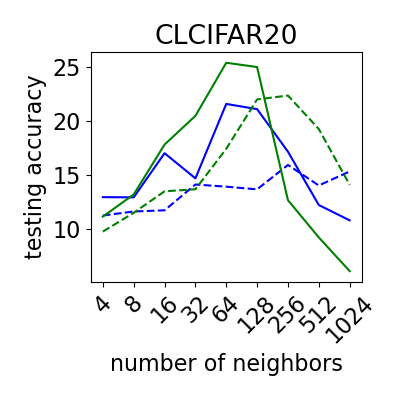}
    }\\
    \vskip-0.1in
    \includegraphics[width=0.5\linewidth]{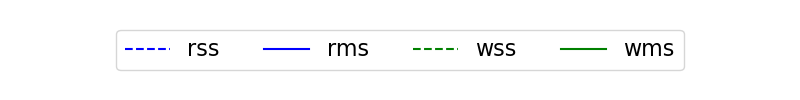}
    \vskip-0.1in
    \caption{
    Comparison of the testing accuracies with different number of neighbors.
    }
    \label{fig:vary-k}
\end{figure}

In this subsection, we discuss how the number of neighbors, the hyperparameter $K$ in Algorithm~\ref{alg:algo}, affects the testing accuracy.
To do so, for each baseline method, we fix the training hyperparameters, including the learning rate and weight decay, to the best one in Section~\ref{sec:experiments} and vary the number of neighbors only.
The results are reported in Figure~\ref{fig:vary-k}.
From the figure, we first observe that the best number of neighbors for the multi-step augmentation is smaller than the best number of neighbors for the single-step augmentation.
The results distinguished the label sharing mechanism between the single-step and multi-step augmentation. The former propagates the label information to distant instances by performing augmentation \emph{multiple} times, whereas the single-step augmentation propagates the label information to more instances by \emph{increasing} the number of neighbors.
Hence, the multi-step augmentation requires less number of neighbors while the single-step augmentation requires more.
Despite the single-step augmentation can be improved by augmenting with more neighboring instances, we still observe that the multi-step augmentation outperforms the single-step augmentation if the number of neighbors can be tuned.

\subsection{Comparison with Alternative Baselines}
\label{subsec:alternative-baselines}
Besides the proposed complementary-label augmentation, there are some alternative ways to utilize the information from the feature space:
\begin{enumerate}
    \item \textbf{Fine-tuning from a self-supervised model} with the loss-based CLL algorithms is also a way to utilize the feature space of the self-supervised models.
    Fine-tuning from a pretrained model can be faster or have better accuracy than training from scratch for some downstream tasks.
    It could potentially be useful for CLL as well.
    \item \textbf{Using a $k$NN method} to obtain a local estimate of the distribution of the complementary labels, then decoding the results by predicting the target label with the least-likely complementary labels.
    This method follows the intuition that the least-likely complementary labels should correspond to the most-likely ordinary labels.
    It corresponds to the decoding methods proposed by \citet{WL2023} when the transition matrix is uniform.
\end{enumerate}
The results are reported in Table~\ref{table:other-baselines-bench}. As we can see from the table, fine-tuning from a self-supervised model improved the testing accuracies of the models in three of the datasets. Nevertheless, label augmentation still improved the fine-tuning method. This implies that label augmentation can provide an orthogonal benefit to fine-tuning. On the other hand, as suggested in the table, $k$NN is a strong baseline for utilizing the information from the self-supervised models, with superior performance to training from scratch and fine-tuning. Still, fine-tuning the self-supervised model with label augmentation outperforms the baseline across all the datasets we consider.

\begin{table}[t]
\caption{
Comparison of the testing accuracy compared with two other baselines: (a) Fine-tuned with pretrained self-supervised models, and (b) $k$ nearest neighbor models. In this table, FT indicates fine-tuning from a self-supervised model, and LA refers to training with complementary label augmentation.
}
\label{table:other-baselines-bench}
\centering
\begin{scriptsize}
\begin{tabular}{lcccc}
\toprule
             & CIFAR10 & CIFAR20 & CLCIFAR10 & CLCIFAR20 \\
\midrule
SCL-NL       &73.95 $\pm$ 0.86 &23.24 $\pm$ 1.54 &44.12 $\pm$ 1.49 &8.09 $\pm$ 0.13 \\
SCL-NL+FT    &84.07 $\pm$ 0.16 &15.31 $\pm$ 2.46 &53.10 $\pm$ 0.74 &10.94 $\pm$ 1.59 \\
SCL-NL+FT+LA &\textbf{89.21 $\pm$ 0.20} &\textbf{66.19 $\pm$ 0.59} &\textbf{78.68 $\pm$ 0.50} &\textbf{29.36 $\pm$ 1.31} \\
\midrule
$k$NN        &86.77 $\pm$ 0.14 &53.12 $\pm$ 0.80 &75.65 $\pm$ 0.41 &27.59 $\pm$ 1.12 \\
\bottomrule
\end{tabular}
\end{scriptsize}
\end{table}

\section{Conclusion and Future Works}
In this paper, we first observe a strong linear correlation between the implicit sharing efficiency and the performance of the CLL algorithms.
This relationship is strong in the sense that it appears for different network architectures, datasets, optimizers, and loss functions.
Based on the observation, we propose \emph{complementary-label augmentation}, a technique that explicitly shares the labeling information between neighboring instances.
The proposed method is compatible with and can be synergistic with the previous CLL algorithms.
Empirical experiments confirm that the proposed method enhances the implicit sharing efficiency and leads to improved performance for previous CLL algorithms on both synthetic and real-world datasets.

The near-linear relationship between the implicit sharing efficiency and the training accuracy leads us to conjecture that there is a theoretical foundation that bridges these two properties.
On the other hand, the proposed methods are limited by the fact that it improves the sharing efficiency in an indirect way.
We leave it to future work to discover the theoretical connection between the implicit sharing efficiency and the training accuracy, and a way to directly enhance the sharing efficiency.

\bibliography{main}
\bibliographystyle{abbrvnat}

\newpage
\appendix
\section{Implementation Details of the Experiments}

\subsection{Implementation Details of the experiments in Section~\ref{sec:methods}}
\label{subsec:appendix-toy-details}

\paragraph{Implicit Label Sharing Experiment}
In this experiment, we trained the model with SCL-NL \cite{chou2020unbiased} loss using the AdamW optimizer with learning rate of $10^{-3}$, weight decay of $10^{-5}$, and batch size of 256 for 100 epochs.
No learning rate scheduling and data augmentation was applied during the training. Three datasets, MNIST\cite{lecun1998gradient}, KuzushijiMNIST\cite{clanuwat2018deep}, and CIFAR10\cite{krizhevsky2009learning} were utilized in this experiment.
The network architectures were a one-layer MLP (784-256-10) for MNIST, LeNet \cite{lecun1998gradient} for Kuzushiji-MNIST, and ResNet-18 \cite{he2016deep} for CIFAR10.
Each trial was repeated five times, and the mean results with standard deviations are reported.
All the experiments in Section~\ref{sec:methods} were run with NVIDIA RTX 2070.

\paragraph{Implicit Label Sharing Experiment with varying parameters}
In this experiment, all the settings are the same as the above except
\begin{enumerate}
    \item Model architectures: we also performed the experiments using one-layer MLP for Kuzushiji-MNIST and LeNet for CIFAR10.
    \item Dataset sizes: we also performed the experiments using only a subset of the dataset. For MNIST and Kuzushiji-MNIST, we additionally trained with size of 40000, 20000, 10000, 8000, 6000, 4000, 2000, 1000, 800, and 600 examples. For CIFAR10, we additionally trained with size of 25000, 10000, 5000, 2500, and 1000 examples.
    \item Optimizers: we also performed the experiments using the SGD optimizer with a momentum of $0.9$, learning rate of $10^{-1}$ and weight decay of $10^{-4}$.
\end{enumerate}
Each trial was repeated five times, and the mean results are reported. Varying these different training settings produces the figures in Figure~\ref{fig:toy-2}.

\paragraph{Explicit Label Sharing Experiment}
The training settings in this experiment are the same as the settings in the implicit label sharing experiment.

\subsection{Implementation Details of the Experiments in Section~\ref{sec:experiments}}
\label{subsec:appendix-exp-details}

We used ResNet-18 as the base model, and trained the SimSiam \cite{chen2021exploring} as the feature extractor to find the neighboring instances.
For SimSiam, we followed the suggested training setting and model architecture (ResNet-18) in the paper.
CLL algorithms were trained using the AdamW optimizer for 200 epochs.
The learning rate was warmed up linearly for the first five epochs, then decayed with cosine annealling for the rest epochs.
The number of neighbors in the proposed label augmentation was fixed to 64, and we set the parameter $\alpha$ to 0.1.
We used standard data augmentation techniques, \texttt{RandomHorizontalFlip}, \texttt{RandomCrop}, and normalization for all training images.
10\% of the data instances in the training datasets were left out as the validation datasets.
The learning rate was selected from $\{10^{-3}, 10^{-4}, 10^{-5}\}$ while the weight decay was selected from $\{10^{-4}, 10^{-5}\}$ using URE on the 0-1 loss on the validation dataset.
It is worth mentioning that the validation dataset consists of only complementary labels.
This protocol is different from some previous works, where the validation dataset consists of ordinary labels.
We argue that our protocol is more practical because ordinary labels may not be obtainable or costly to collect in real world.

For the experiments in Section~\ref{subsec:alternative-baselines}, the training settings were the same as above, except that the number of epochs was reduced to 50. The hyperparameter $k$ for the $k$NN was selected from $\{4,8,16,32,64,128,256,512,1024\}$.

We completed five trials for each experiment with NVIDIA V100.

\section{Additional Results of the Experiments}

\subsection{Additional Results of the Experiments in Section~\ref{sec:methods}}
\label{subsec:appendix-toy-additional-results}
\begin{figure}[t]
    \centering
    \resizebox{\columnwidth}{!}{
    \includegraphics[width=0.3\linewidth]{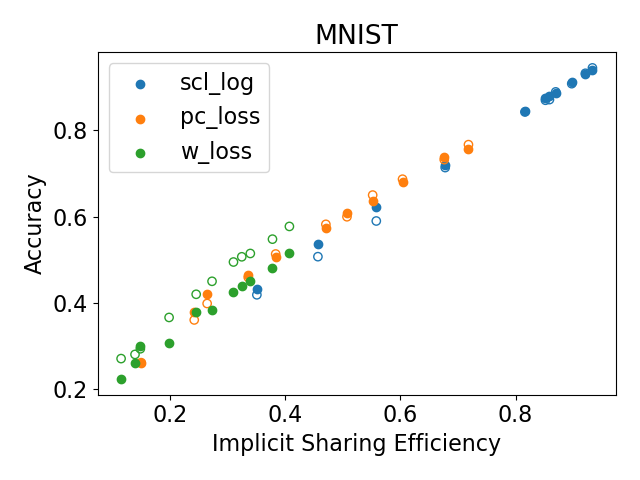}
    \includegraphics[width=0.3\linewidth]{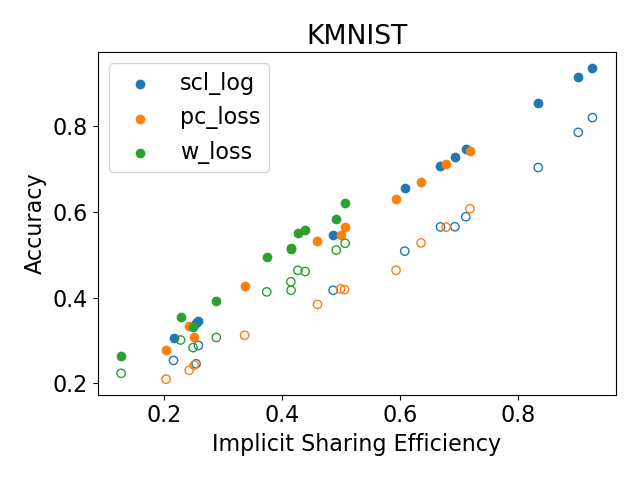}
    \includegraphics[width=0.3\linewidth]{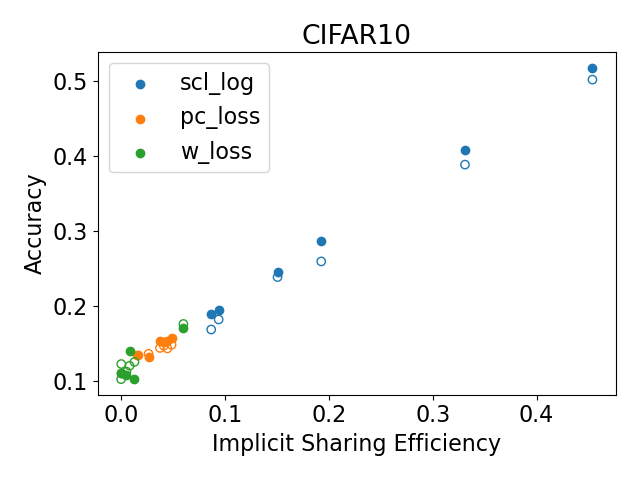}
    }
    \caption{
        Relationship between the implicit sharing efficiency and model performance.
        Each dot represents a result of the last epoch in training with a certain CLL algorithm.
        Specifically, solid markers ($\bullet$) refer to the training accuracy while empty markers ($\circ$) refer to the testing accuracy.
    }
    \label{fig:method-eff-acc}
\end{figure}

In this section, we repeated the experiments with two additional CLL algorithms:
(a) PC \cite{ishida2017learning}: the pairwise comparison loss,
(b) L-W \cite{gao2021discriminative}: the weighted loss derived from the discriminative modeling of the complementary labels.
The results between the implicit sharing efficiency and model performance are reported in Figure~\ref{fig:method-eff-acc}.
As we can observe, the positive correlation between those two properties still hold even if we train the model with different methods.

\subsection{Additional Results of the Experiments in Section~\ref{sec:experiments}}
In the standard benchmark in Section~\ref{sec:experiments}, we assessed the models' performance using the model in the last epoch.
As the CLL algorithms are known to have a tendency to overfit, some previous studies selected the best epoch using the validation dataset instead of selecting the last epoch.
For instance, \citet{ishida2019complementary} selected the best epoch on the validation dataset while \citet{yu2018learning} employed early-stopping.
For completeness, we report the results on the standard benchmark where we selected the model based on the best epoch on the validation dataset using the URE of the 0-1 loss.
The results are reported in Table~\ref{table:clean-bench-earlystop}.
As we can observe, the proposed label augmentation still improves the previous methods in this training setting, demonstrating the broad applicability of the proposed method.

\begin{table}[t]
\caption{
Effects of label augmentation applied on different complementary learning algorithms on CIFAR10 and CIFAR20, where the best epoch selected using the validation dataset is evaluated. The mean and standard deviation of the testing accuracies over five random trials are reported.
}
\label{table:clean-bench-earlystop}
\vskip 0.15in
\centering
\begin{footnotesize}
\begin{tabular}{lcccc}
\toprule
        & CIFAR10 & CIFAR20 & CLCIFAR10 & CLCIFAR20 \\
\midrule
PC     &40.11 $\pm$ 2.36 &12.89 $\pm$ 1.82 &44.32 $\pm$ 1.69 &15.20 $\pm$ 1.54 \\
PC+RSS &72.52 $\pm$ 0.80 &33.53 $\pm$ 0.73 &66.12 $\pm$ 1.48 &26.25 $\pm$ 1.13 \\
PC+RMS &83.27 $\pm$ 0.52 &53.29 $\pm$ 1.14 &73.52 $\pm$ 1.35 &26.15 $\pm$ 2.40 \\
PC+DSS &80.99 $\pm$ 0.27 &43.74 $\pm$ 0.46 &72.67 $\pm$ 0.77 &26.80 $\pm$ 3.12 \\
PC+DMS &\textbf{83.61 $\pm$ 0.59} &\textbf{55.59 $\pm$ 1.58} &\textbf{74.09 $\pm$ 1.59} &\textbf{30.94 $\pm$ 1.89} \\
\midrule
URE-GA     &56.88 $\pm$ 1.59 &11.41 $\pm$ 1.47 &25.84 $\pm$ 1.12 &9.12 $\pm$ 0.39 \\
URE-GA+RSS &72.58 $\pm$ 1.42 &23.81 $\pm$ 3.18 &\textbf{28.36 $\pm$ 1.00} &11.37 $\pm$ 0.47 \\
URE-GA+RMS &72.91 $\pm$ 2.59 &24.43 $\pm$ 1.42 &24.52 $\pm$ 2.64 &12.11 $\pm$ 0.43 \\
URE-GA+DSS &73.32 $\pm$ 1.75 &25.64 $\pm$ 2.69 &26.35 $\pm$ 1.30 &12.50 $\pm$ 0.47 \\
URE-GA+DMS &\textbf{75.40 $\pm$ 2.39} &\textbf{27.64 $\pm$ 3.43} &20.44 $\pm$ 1.66 &\textbf{13.01 $\pm$ 0.49} \\
\midrule
SCL-NL     &72.80 $\pm$ 1.21 &22.37 $\pm$ 1.77 &48.38 $\pm$ 2.18 &8.08 $\pm$ 0.51 \\
SCL-NL+RSS &86.35 $\pm$ 0.12 &56.70 $\pm$ 0.82 &65.53 $\pm$ 1.59 &13.09 $\pm$ 1.33 \\
SCL-NL+RMS &88.35 $\pm$ 0.22 &\textbf{63.87 $\pm$ 1.09} &74.17 $\pm$ 2.66 &19.84 $\pm$ 1.16 \\
SCL-NL+DSS &86.91 $\pm$ 0.27 &56.79 $\pm$ 1.35 &70.11 $\pm$ 1.55 &18.58 $\pm$ 1.92 \\
SCL-NL+DMS &\textbf{88.49 $\pm$ 0.37} &63.85 $\pm$ 2.73 &\textbf{77.80 $\pm$ 1.87} &\textbf{23.49 $\pm$ 1.93} \\
\midrule
L-W     &64.82 $\pm$ 0.39 &16.20 $\pm$ 1.69 &47.22 $\pm$ 2.09 &7.98 $\pm$ 0.41 \\
L-W+RSS &85.46 $\pm$ 0.37 &50.11 $\pm$ 0.59 &65.26 $\pm$ 1.58 &15.01 $\pm$ 1.09 \\
L-W+RMS &87.70 $\pm$ 0.18 &62.76 $\pm$ 0.69 &73.49 $\pm$ 2.58 &19.52 $\pm$ 2.52 \\
L-W+DSS &86.44 $\pm$ 0.24 &54.50 $\pm$ 1.23 &71.48 $\pm$ 2.06 &18.92 $\pm$ 1.10 \\
L-W+DMS &\textbf{87.83 $\pm$ 0.47} &\textbf{65.08 $\pm$ 0.70} &\textbf{77.46 $\pm$ 1.46} &\textbf{23.52 $\pm$ 1.42} \\
\bottomrule
\end{tabular}
\end{footnotesize}
\end{table}

\end{document}